\documentclass[letterpaper]{article} 
\usepackage{aaai23}  
\usepackage{times}  
\usepackage{helvet}  
\usepackage{courier}  
\usepackage[hyphens]{url}  
\usepackage{graphicx} 
\usepackage{natbib}  
\usepackage{caption} 
\usepackage{algorithm}
\usepackage{algorithmic}
\usepackage{latexsym}

\usepackage{color}
\usepackage{graphicx}
\usepackage{amsmath}
\usepackage{amssymb}
\usepackage{amsfonts}
\usepackage{booktabs}
\usepackage{multirow}
\usepackage{bm}
\usepackage{newfloat}
\usepackage{listings}
\DeclareCaptionStyle{ruled}{labelfont=normalfont,labelsep=colon,strut=off} 
\floatstyle{ruled}
\newfloat{listing}{tb}{lst}{}
\floatname{listing}{Listing}
\title{Exploring Faithful Rationale for Multi-hop Fact Verification via Salience-Aware Graph Learning}
\author{
    Jiasheng Si,
    Yingjie Zhu, 
    Deyu Zhou\thanks{Corresponding author.}
}
\affiliations{
    School of Computer Science and Engineering, Key Laboratory of Computer Network \\ and Information Integration, Ministry of Education, Southeast University, China\\

    \{jasenchn, yj\_zhu, d.zhou\}@seu.edu.cn
}

\usepackage{bibentry}

\begin{document}

\maketitle

\begin{abstract}
    The opaqueness of the multi-hop fact verification model imposes imperative requirements for explainability. One feasible way is to extract \textit{rationales}, a subset of inputs, where the performance of prediction drops dramatically when being removed. Though being explainable, most rationale extraction methods for multi-hop fact verification explore the semantic information within each piece of evidence individually,
    while ignoring the topological information interaction among different pieces of evidence. 
    Intuitively, a faithful rationale bears complementary information being able to extract other rationales through the multi-hop reasoning process. To tackle such disadvantages, we cast explainable multi-hop fact verification as subgraph extraction, which can be solved based on graph convolutional network (GCN) with salience-aware graph learning. 
    In specific, GCN is utilized to incorporate the topological interaction information among multiple pieces of evidence for learning evidence representation. 
    Meanwhile, to alleviate the influence of noisy evidence, the salience-aware graph perturbation is induced into the message passing of GCN. 
    Moreover, the multi-task model with three diagnostic properties of rationale is elaborately designed to improve the quality of an explanation without any explicit annotations. 
    Experimental results on the FEVEROUS benchmark show significant gains over previous state-of-the-art methods for both rationale extraction and fact verification.
\end{abstract}

\section{Introduction}
\label{sec:introduction}
The wide availability of user-provided content on online social media facilitates the rapid dissemination of unfounded rumors and misinformation. 
Fact verification, automatically assessing the veracity of a textual claim against multiple pieces of evidence retrieved from external sources, has gained intense attention to combat misinformation spread on the internet~\citep{DBLP:journals/csur/ZhouZ20,DBLP:journals/corr/abs-2108-11896,DBLP:conf/acl/SiZLSH20}. 
However, as the opaqueness of the model diminishes user confidence and impedes the discovery of harmful biases~\citep{DBLP:conf/coling/KotonyaT20}, 
it is essential to understand the ``reasoning'' behind the model prediction, i.e, the explainability of the fact verification approaches.

\begin{figure}[t]
  \centering
  \includegraphics[scale=0.53]{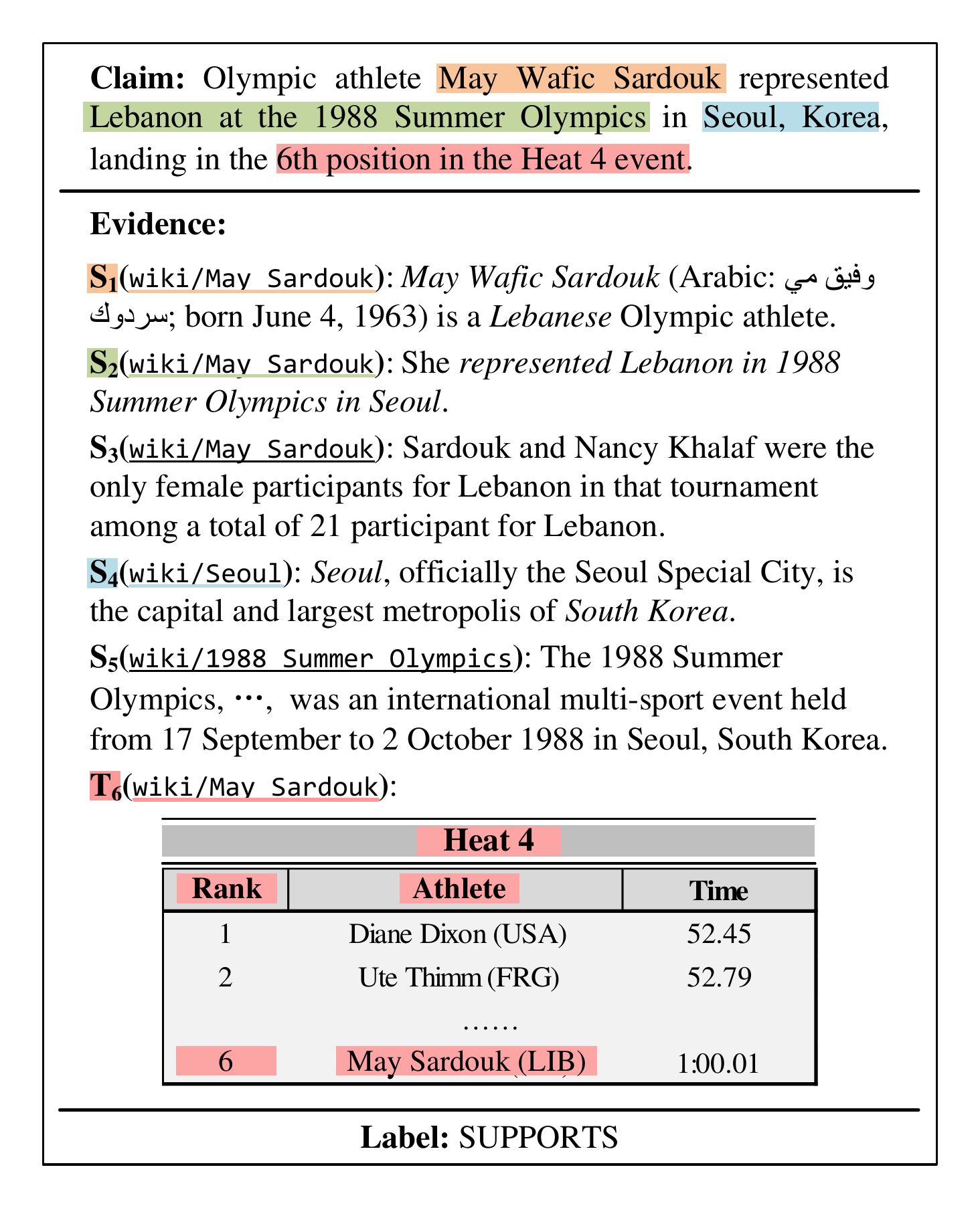}
  \caption{An example from FEVEROUS dataset, where \textit{S1, S2, S4} and two table cells in \textit{T6} are considered as rationales.}
  \label{fig:example}
\end{figure}

A straightforward way to generate explanations for fact verification is to use \textbf{rationale extraction}~\citep{zaidan2007using, DBLP:conf/acl/DeYoungJRLXSW20,DBLP:conf/acl/AtanasovaSLA20,DBLP:conf/emnlp/GlocknerHG20,DBLP:journals/corr/abs-2109-03756},
a \textit{post-hoc technique} searching for a minimal portion of input (i.e., rationales) that can be sufficient (i.e., solely based on the rationales) to derive a veracity prediction. 
The intuition is that the retrieved evidence for verifying the claim comprises noisy evidence inevitably, and the interaction of true evidence\footnote{Note that we make a distinction between \textit{true evidence} and \textit{noisy evidence} conceptually, and define the \textit{true evidence} as the \textit{rationale} since the term ``rationale'' implies human-like intent~\citep{DBLP:journals/corr/abs-2102-12060}.} is adequate for the multi-hop fact verification model to reach a disposition accurately. 
It contrasts with the methods heuristically exploring the importance of input features, such as attention-based methods~\citep{chen2020loren,DBLP:conf/acl/WuRL0Q20} or gradient-based methods~\citep{DBLP:conf/icml/SundararajanTY17},
which inevitably induce low-scoring features, drawing criticism recently~\citep{DBLP:conf/naacl/JainW19}.
In this work, we focus on how to extract valid rationales for explaining multi-hop fact verification model. 

Existing rationale extraction methods for multi-hop fact verification model usually rely on the FEVER dataset~\citep{DBLP:conf/naacl/ThorneVCM18,DBLP:conf/acl/DeYoungJRLXSW20,bekoulis2021review}.
These methods typically decompose the model into the extractor module and the predictor module independently~\citep{DBLP:conf/emnlp/LeiBJ16,DBLP:conf/acl/JainWPW20,DBLP:conf/emnlp/KotonyaT20}, 
where the former is trained to assign a mask score over the subset (e.g., sentences or tokens) of inputs to capture the rationale, 
then the latter makes predictions exclusively on the rationales provided by the extractor.
The quality of the explanation depends upon the strategy of the mask vector training.

However, despite the salient progress, there are still limitations required to be addressed for explainable multi-hop fact verification~\citep{DBLP:conf/emnlp/JiangBZD0B20,DBLP:conf/ijcai/OstrowskiAAA21,DBLP:conf/nips/AlyGST00CM21,DBLP:conf/acl/AtanasovaSLA20,DBLP:conf/acl/JainWPW20,DBLP:journals/corr/abs-2109-03756}. 
Inherently, in multi-hop fact verification, 
the claims may be verified by aggregating and reasoning over multiple pieces of rationales.
For example in Fig.\ref{fig:example}, the truthfulness of the claim can be assessed by aggregating four pieces of true evidence (i.e., rationales), including sentences and table cells, surrounded by multiple pieces of noisy evidence.
Intuitively, the rationale \textit{S1} carries the complementary information capable of extracting the rationale \textit{S2} through the multi-hop reasoning process.
However, in the mask vector learning process, 
current rationale extraction methods are mainly based on the semantic information of individual semantic units (sentence or token) within the input, 
failing to capture the topological information interaction among multiple pieces of the semantic unit in the multi-hop reasoning process for rationale extraction, which we argue is crucial for the explainable multi-hop fact verification.

To address such disadvantage, we introduce a GCN-based model~\citep{DBLP:conf/iclr/KipfW17} with \textbf{Sa}lience-aware \textbf{G}raph \textbf{P}erturbation, namely SaGP, 
where \textit{multi-hop fact verification} and sentence-level \textit{rationale extraction} are optimized jointly.
The \textbf{core novelty} here is that we frame the rationale extraction of multi-hop fact verification as \textit{subgraph extraction} 
via searching for the rationale subgraph with minimal nodes while sufficiently maintaining the prediction accuracy.
Specifically, we use GCN to integrate the topological interaction of information among different pieces of evidence to update evidence representation.
Meanwhile, to alleviate the influence of the noisy evidence in this process, we induce a learnable salience-aware perturbation (edge mask, node mask) into the message passing process of GCN to approximate the deletion of the superfluous edges or nodes in the input graph. It guarantees that the information masked out from the graph is not propagated for evidence representation learning. 
Then the assignment vector over each node is learned to indicate whether the evidence could be contained in the rationale subgraph,
which approximates the mask vector learning following prior works.
Moreover, we incorporate the multi-task learning paradigm and define three diagnostic properties (i.e., \textit{Fidelity}, \textit{Compact}, \textit{Topology}) as additional optimizing signals to guide the learning of rationale subgraph.

The main contributions are listed as follows:
(I) We frame the explainable multi-hop fact verification as subgraph extraction, where a GCN-based model with salience-aware graph learning is proposed.
(II) The multi-task model with three diagnostic properties is designed and optimized to improve the quality of extracted explanations without accessing the rationale supervision.
(III) Experimental results on the FEVEROUS dataset show the  superior performance of the proposed approach.

\section{Related Works}
\label{related_works}

Fact verification is the task of assessing the veracity of the claim backed by multiple pieces of validated evidence, 
which can be decomposed into two stages: evidence retrieval and claim verification~\citep{DBLP:conf/acl/SiZLSH20}.
Thus, two aspects can be explored for explaining the fact verification model: 
(I) retrieving faithful evidence as precisely as possible in the evidence retrieval stage~\citep{DBLP:conf/acl/WanCDLY20};
(II) extracting rationales with the explainable techniques in the claim verification stage. 
Researchers mainly focus on the latter by developing explainability techniques for fact verification, which broadly fall into the scope of attention-based methods by treating the attention weights over input representation as a measure of credibility score~\citep{DBLP:conf/coling/KotonyaT20,DBLP:journals/corr/abs-2103-11072}.
Attention mechanisms vary in the information type of input, including self-attention~\citep{DBLP:conf/emnlp/PopatMYW18}, 
co-attention~\citep{DBLP:conf/kdd/ShuCW0L19,DBLP:conf/acl/WuRL0Q20,DBLP:conf/www/YangPMDYLRJH19,DBLP:conf/acl/WuRZLN20}.
It provides an optional way for multi-hop fact verification,
while recently some works argued that attention weights cannot guarantee the inattention of low-confidence features, and are not valid explanations for model prediction~\citep{DBLP:conf/naacl/JainW19,DBLP:conf/acl/SerranoS19,DBLP:conf/acl/MeisterLAC20}.

Other researches focus on exploring the rationale extraction for explaining the fact verification via perturbing the input~\citep{DBLP:journals/corr/abs-2103-11072,DBLP:conf/coling/KotonyaT20},
which usually consists of two modules, i.e., extractor and predictor~\citep{DBLP:conf/acl/AtanasovaSLA20,DBLP:conf/emnlp/ParanjapeJTHZ20,DBLP:conf/aaai/ShaCL21}. 
Generally, the predictor is used to devise the decision based on the rationales generated from the extractor rather than the whole input.
However, it is tricky to jointly train the two separate modules because of the intractable rationale sampling. It is partially solved via reinforcement learning~\citep{DBLP:conf/emnlp/LeiBJ16,DBLP:conf/iclr/YoonJS19}, 
or reparameterization techniques~\citep{DBLP:conf/acl/BastingsAT19}.
An alternative way is to adopt multi-task learning~\citep{DBLP:journals/corr/abs-2109-03756,DBLP:journals/corr/abs-2102-12060}, which provides more label-informed rationales than prior methods.
However, 
they mainly extract the rationales with the semantic information of individual semantic units of the input,
which is not able to capture the topological information interaction among different pieces of evidence in the mask vector learning process. 
It may not be suitable to explain multi-hop fact verification~\citep{DBLP:conf/emnlp/GlocknerHG20}.

Different from the approaches mentioned above, our work is the first to formulate the explainable multi-hop fact verification as subgraph extraction. 
Moreover, we utilize three properties to guide the extraction of the rationale subgraph in multi-task learning.

\section{Methodology}
\label{sec:methodology}

\subsection{Problem Setting}
Assume that a claim may be verified with multi-hop evidence, including sentences, table cells, or a combination of multiple sentences or table cells~\cite{DBLP:conf/nips/AlyGST00CM21}. Given a claim $\bm{c}$ with associated textual evidence (i.e., sentences, table captions) $\{\bm{t}_1,\bm{t}_2,...,\bm{t}_S\}$ or tabular evidence (i.e., table cells) $\{\bm{c}_{1},\bm{c}_2,...,\bm{c}_C\}$, 
a heterogeneous evidence graph $G=(X, A)$ is constructed to model how the claim is associated with the textual evidence and the tabular evidence. 
Each node $x_i \in X$ represents an evidence sequence by concatenating the claim and the textual evidence (i.e., sentences or table captions) or tabular evidence (i.e., cell sequences). 
$A$ denotes an adjacency matrix for the undirected fully connected graph with the edge weight equal to 1. 
Given a trained GNN-based multi-hop fact verification model,
we aim to extract the rationale subgraph (\textit{RA-subgraph}) for this model by incorporating the topological information interaction among different pieces of evidence. Nodes within the extracted subgraph are treated as the rationale. 
Therefore, the aim of explainable multi-hop fact verification is to infer the claim verification label $\hat{y}^c$ as \textit{SUPPORTS} or \textit{REFUTES} and to assign each sentence or table cell with a Boolean label denoted as $\hat{y}^e_i\in\{0, 1\}$, where $\hat{y}^e_i=1$ denotes that the sequence $i$ is in the RA-subgraph which actually benefits the predictions.

\subsection{The Architecture}
\label{subsection:architecture}
\begin{figure*}[]
  \centering
  \includegraphics[scale=0.6]{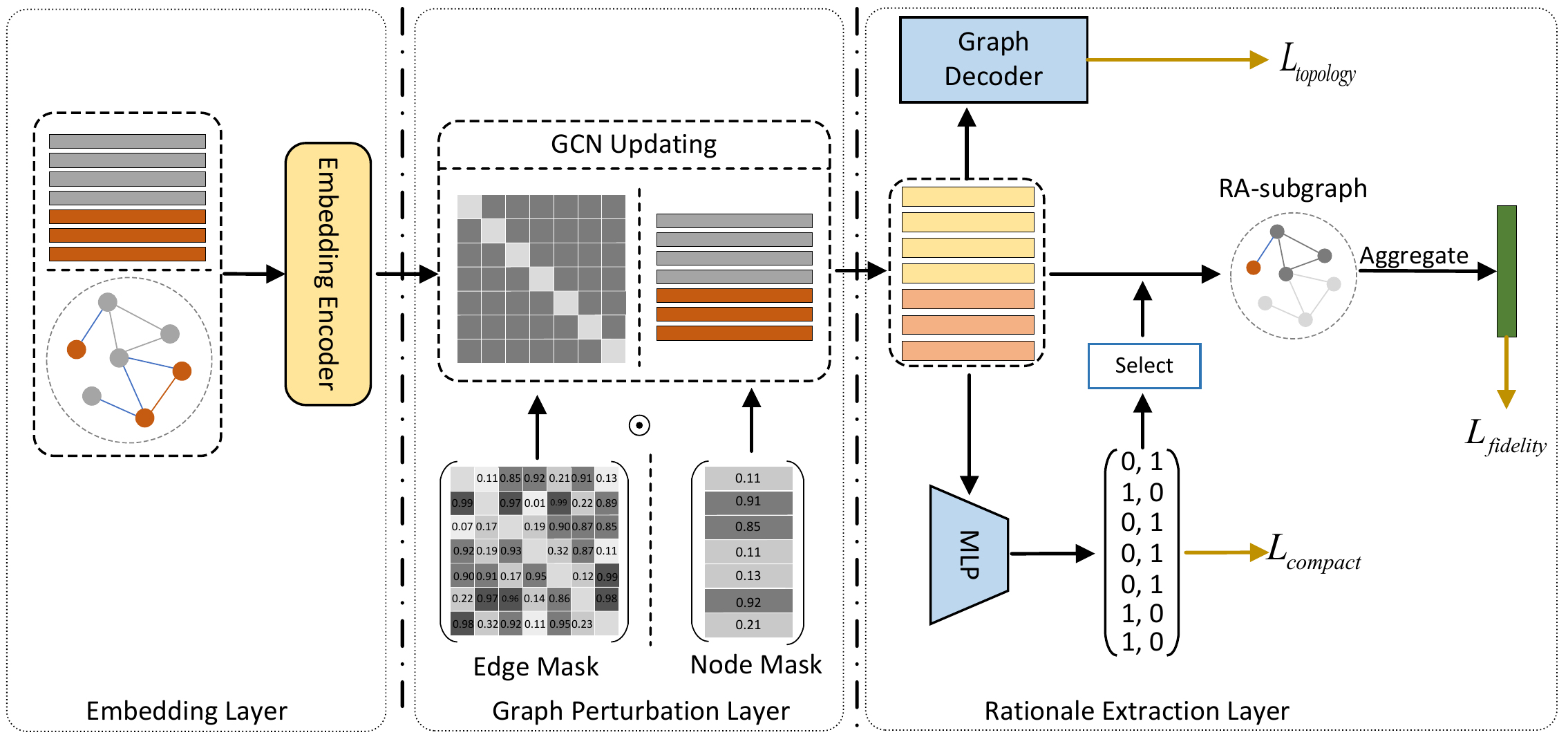}
  \caption{The overall framework of the proposed SaGP.}
  \label{fig:model_architecture}
\end{figure*}

We propose the method~\textbf{SaGP}, 
which consists of three key components as shown in Fig.~\ref{fig:model_architecture}: 
(I) an embedding layer, where a pre-trained language model is employed to obtain the initial node representations of the input graph.
(II) a graph perturbation layer, where evidence representation is updated by inducing the salience-aware graph perturbation into the message passing of GCN.
(III) a rationale extraction layer, where the RA-subgraph is extracted by directly optimizing for three diagnostic properties (i.e., Fidelity, Compact, Topology) of rationales in the multi-task model. 
The details of each component are provided in the following sections.

\subsection{Embedding Layer}
\label{subsubsec:embedding_layer}
It is challenging to manipulate the tabular evidence at cell-level. 
Following~\citet{DBLP:conf/iclr/ChenWCZWLZW20,DBLP:journals/corr/abs-2109-12349}, we employ the simple table linearization template to generate contextualized per-cell sequence representations to form the cell sequence of the table, where each table cell is linearized as ``\emph{In \underline{wikipage}, the header is \underline{tableheader1} (and \underline{tableheader2}), the value is \underline{tablecell}.}''.
Different from TAPAS~\citep{DBLP:conf/acl/HerzigNMPE20}, we do not consider the table structure since there are some noisy cells contained in the table, which might confuse the model. It should be pointed out that in our experiments,  there is no substantial improvement when employing elaborated templates such as~\citet{DBLP:journals/corr/abs-2109-12349}.

For the initial node representation in the input evidence graph, a RoBERTa~\cite{DBLP:journals/corr/abs-1907-11692} model is employed to encode each evidence sequence $x_i$ to its contextual embedding $\bm{h}_i=encoder(x_i)$ by selecting the hidden representation of the \emph{CLS} token from the final layer.

\subsection{Graph Perturbation Layer}
\label{sec:graph_perturbation_layer}

To identify rationales by incorporating the topological interaction of different pieces of evidence while mitigating the influence of noisy evidence,
we induce a learnable salience-aware graph perturbation into the messaging passing of GCN.
Specifically, given a trained GCN-based model, 
taking an input graph as input, 
we probe the GCN layer using a learnable salience-aware graph perturbation matrix (a.k.a., edge mask or node  mask)~\citep{DBLP:conf/nips/YingBYZL19}, where the gates in the perturbation matrix indicate which edges or nodes are necessary and which can be disregarded. 
The assumption behind this is that the nodes within the neighborhood might contain noise or even conflicting information,
while the \textit{zero} edges or nodes bear no information and can be removed from the message aggregation for node updating,
which is analogous to the perturbation of input.
The masking may also be equivalently seen as adding a certain type of noise to the message passing process in the GCN.

Let $f$ be a trained GCN layer for node representation learning,
\begin{equation}
  \label{equ:gcn}
  f(A,H;W) = relu(\widetilde{D}^{-1/2}\widetilde{A}\widetilde{D}^{-1/2}HW),
\end{equation}
where $\widetilde{A}=A+I$, 
$I$ is the identity matrix,
$\widetilde{D}$ is the degree matrix,
$H$ denotes the evidence embeddings and 
$W$ denotes the parameters of GCN.

Given the trained GCN layer with parameters $W$ frozen, we seek to introduce the learnable parameters to mitigate the impact of noise evidence in the GCN message passing process.
For the edge mask, we introduce a learnable perturbation matrix $P$ with the same size as the adjacency matrix $A$ to approximate zero out the superfluous entries in the adjacency matrix. 
Each element $p_{ij}$ in matrix $P$ indicates the importance score for message aggregation from node $i$ to node $j$, 
where the sigmoid transformation is utilized to restrict the matrix with entries in $[0, 1]$.
Specifically, if element $p_{ij}\rightarrow 0$, it results in the deletion of the edge from node $i$ to node $j$.
We populate $P$ in an asymmetrical manner since we argue the information discrepancy for different nodes within a node pair. 
In other words, the node $i$ is useful for updating the node $j$ cannot guarantee that the opposite is valid.
The calculation in the GCN can be rewritten into

\begin{small}
\begin{equation}
  \label{equ:gcn_edge_mask}
  \widetilde{f}(A,H;W) = relu(((\widetilde{D}^{-1/2}\widetilde{A}\widetilde{D}^{-1/2})\odot \sigma{(P)})HW),
\end{equation}
\end{small}
where $\odot$ denotes the element product.

For node mask, similarly, we introduce a learnable perturbation matrix $M$ to assign the attribution score for each node with the sigmoid transformation, which indicates the degree of node features used in the message passing in the graph.

\begin{small}
\begin{equation}
  \label{equ:gcn_node_mask}
  \widetilde{f}(A,H;W) = relu(\widetilde{D}^{-1/2}\widetilde{A}\widetilde{D}^{-1/2}(H\odot \sigma{(M)})W)
\end{equation}
\end{small}

There are no supervision signals on the graph to update the perturbation matrix of the probe.
Therefore, two regulations are applied to avoid all entries in the matrix approximating 1: 
(I) the sum regulation $\mathcal{L}_{sum}$ of all entries in the perturbation matrix to constrain the size of the perturbation;
(II) the information entropy regulation $\mathcal{L}_{entropy}$ to reduce the uncertainty of the perturbation matrix.

\subsection{Rationale Extraction Layer}
\label{sec:rationale_extraction_layer}

Connecting the explanation extractor module and predictor module with the learnable mask vector is indispensable for valid rationale extraction.
Previous methods generally devise complicated strategies for linking the two modules (e.g., REINFORCE-based~\citep{DBLP:conf/emnlp/LeiBJ16}, heuristic-based~\citep{DBLP:conf/acl/JainWPW20}), 
which is inefficient.
Following~\citet{DBLP:journals/corr/abs-2109-03756}, 
we employ the multi-task learning paradigm in which explainable multi-hop fact verification is factorized into first extracting the RA-subgraph from the input graph and then conducting the claim verification conditioned on RA-subgraph. 
Moreover, three diagnostic properties of the rationale subgraph are optimized jointly.

\subsubsection{Rationale Subgraph Extraction}
\label{subsubsec:rationale_subgraph_extraction}

Given the node representation $\widetilde{U}\in\mathbb{R}^{N\times d}$ learned by the \textbf{perturbed GCN} $\widetilde{f}$ (in Sec.~\ref{sec:graph_perturbation_layer}), 
we first extract the subgraph via the node assignment $S\in\mathbb{R}^{N\times2}$,
where the first dimension indicates the probability of nodes is the rationale and could be included in the rationale subgraph $G_{sub}$~\citep{DBLP:conf/iclr/YuXRBHH21}, 
and the second dimension denotes the non-rationale and the complementary subgraph of $G_{sub}$ in input graph, namely $\overline{G}_{sub}$.
\begin{equation}
  \label{equ:subgraph}
  S = softmax(MLP(\widetilde{U};W_{sub})),
\end{equation}
where $MLP$ is a linear layer, $W_{sub}$ is the learnable parameters.
Then, the representation $R_{sub}$ of the RA-subgraph $G_{sub}$ can be aggregated by taking the first column of $S$, i.e., $R_{sub}=S_1^T\widetilde{U}$.
Consequently, a linear layer with softmax operation on $R_{sub}$ predicts the target label $\hat{y}_{sub}^{c}\in\mathbb{R}^2$.

\subsubsection{Property}
We propose three diagnostic properties to regularize the rationale subgraph extraction.

\textbf{Fidelity.}
The faithful RA-subgraph extracted from an input graph ought to exhibit meaningful influence on its prediction.
Thus, we define the cross entropy function to measure the \textit{fidelity} for faithfulness of RA-subgraph~\citep{DBLP:conf/acl/DeYoungJRLXSW20}.
In particular, given the node representation $U$ learned from the \textbf{standard GCN} $f$, a linear layer with softmax operation is employed to map the representation of the full graph to its prediction $\hat{y}^c_{full}$ by aggregating $U$.
We assume that the rationales within the subgraph are adequate to match the original prediction.
\begin{equation}
  \label{equ:faithful}
  L_{fidelity} = CrossEntropy(\hat{y}^c_{sub}(\widetilde{U}), \hat{y}^c_{full}(U)),
\end{equation}
intuitively, a low score here implies that the rationale contained in the subgraph is indeed susceptible on the prediction.

\textbf{Compact.}
The compact measures how distinctive the node assignment within the graph is. Including it as an objective can serve as an additional regularization on $S$ for the model to be compact in the extracted subgraph.
The poor perturbation, on the one hand, will enable $p(x_i)\in G_{sub}$ and $p(x_i)\in \overline{G}_{sub}$ to be close.
Additionally, the model is prone to assigning all the nodes to $p(x_i)\in G_{sub}$, resulting in redundancy for the RA-subgraph.
To do so, we employed the compact loss to distinguish node assignments and urge the RA-subgraph to be compact,
\begin{equation}
  \label{equ:compact}
  L_{compact} = \| norm(S^TAS) - I_2 \|_F,
\end{equation}
where $norm(\cdot)$ denotes the normalization, $\|\cdot\|_F$ denotes the Frobenius norm, $I_2$ is the identity matrix.

\textbf{Topology.}
The topology measures whether the representation learned from the perturbed GCN $\widetilde{f}$ preserves the original topological information. We consider this a valuable objective to re-calibrate due to the asymmetrical information passing in the perturbed graph.
Motivated by graph auto-encoder~\citep{DBLP:journals/corr/KipfW16a},
we define a decoder to reconstruct the original graph adjacency $A$ using the perturbed node representations $\widetilde{U}$, 

\begin{equation}
  \begin{aligned}
  \label{equ:topological_structure}
  L_{topology} &= CrossEntropy(\hat{A}, A), \\
  \hat{A}&=\sigma(\widetilde{U}\widetilde{U}^T)
  \end{aligned}
\end{equation}
where $\sigma$ denotes the sigmoid function.

\subsection{Objective for Learning}
\label{subsec:objective}

The overall objective function $\mathcal{L}$ is minimized over the above modules:
\begin{equation}
  \begin{aligned}
  \label{equ:loss}
  \mathcal{L} =& \lambda_1\mathcal{L}_{fidelity} + \lambda_2\mathcal{L}_{compact} + \lambda_3\mathcal{L}_{topology} \\
  & + \lambda_4\mathcal{L}_{sum} + \lambda_5\mathcal{L}_{entropy},
  \end{aligned}
\end{equation}
where $\lambda_{1-5}$ are hyperparameters.
In addition, we elucidate how the rationale supervision can be used for rationale extraction, 
where the $\mathcal{L}_{compact}$ is replaced by the rationale loss, 
which treats $S$ as a standard multi-label problem via a sigmoid layer and binary cross entropy function. 
This is a compromised way to use supervision at the node-level, as it is difficult to obtain the real edge tag.

\section{Experimental Setup}
\label{sec:experimental_setup}
This section describes the datasets, evaluation metrics and the baselines in the experiments.

\textbf{Dataset.}
We conduct our experiments on the large scale dataset FEVEROUS~\citep{DBLP:conf/nips/AlyGST00CM21}, 
which is a multi-hop dataset with different types of evidence. 
To construct the explanatory dataset with rationale annotation, the claims with the \textit{NOT ENOUGH INFO} label are deleted because there are no gold standard rationales corresponding to that label. 
Moreover, apart from retaining the rationales in the dataset,
we retrieved the complementary evidence relevant to the claim as the \textit{noisy evidence} using the method described in~\citet{DBLP:conf/nips/AlyGST00CM21}.
We construct the dataset where each claim associates with 20 pieces of evidence, including \textit{true evidence} (i.e., rationale) and \textit{noisy evidence}, 
wherein the verdict of the claim requires reasoning over the aggregation of the sentences and table cells (1.43 sentences and 3.42 table cells in the training dataset, 1.43 sentences and 2.83 table cells in the test dataset on average).
The statistics of the dataset are shown in Tab.\ref{tab:statistics}.

\textbf{Metrics.}
We first adopt the metrics proposed for the ERASER benchmark~\citep{DBLP:conf/acl/DeYoungJRLXSW20} to measure the agreement with the human-annotated rationales by evaluating the macro F1, Precision, and Recall metrics,
where we choose the sentences as the basic unit for rationales.
We also report the Exact Match Accuracy (Ext.acc) for strict measuring rationale comparison. 
Secondly, 
We adopt the macro F1 and Accuracy metrics for claim verification evaluation.
In addition,
we report the joint accuracy of the claim verification and the rationale extraction, 
where we consider the prediction is correct if the predictions are correct for the two tasks, 
where
Acc.Full denotes the correct prediction of the claim with all rationales,
Acc.Part denotes the correct prediction of the claim with one piece of rationale.

\begin{table}[t]\small
  \renewcommand\arraystretch{1.1}
  \renewcommand\tabcolsep{2.5pt}
  \centering
  \begin{tabular}{cl|ccccc}
  \hline
  \multicolumn{2}{c|}{\textbf{FEVEROUS}} & \textbf{Num.Sup} & \textbf{Num.Ref} & \textbf{Avg.Ra} & \textbf{Avg.S} & \textbf{Avg.C} \\ \hline
  \multicolumn{2}{c|}{\textbf{Train}}    & 41,835           & 27,215           & 4.85            & 1.43           & 3.42           \\
  \multicolumn{2}{c|}{\textbf{Test}}     & 3,908            & 3,481            & 4.26            & 1.43           & 2.83           \\ \hline
  \end{tabular}
  \caption{\label{tab:statistics} Statistics of the FEVEROUS dataset.
  \textit{Num.Sup} and \textit{Num.Ref} are the number of claims with \textit{SUPPORT} label and \textit{REFUTE} label.
   \textit{Avg.Ra}, \textit{Avg.S}, and \textit{Avg.C} denote the average number of \textit{rationales}, \textit{sentence rationales}, \textit{table cell rationales} per claim, respectively.}
\end{table}

\textbf{Baselines.}
We compare the proposed SaGP model with the following baselines for claim verification and rationale extraction under unsupervised and supervised settings,
including 
(I) the pipeline method from ERASER~\citep{DBLP:conf/acl/DeYoungJRLXSW20}, which verdicts the claim using one rationale with the highest score from the extractor. 
(II) the information bottleneck (IB) method~\citep{DBLP:conf/emnlp/ParanjapeJTHZ20}, which extracts sentence-level rationales by measuring the mutual information with the label.
(III) the two-sentence selecting (TSS) method~\citep{DBLP:conf/emnlp/GlocknerHG20}, which extracts the rationales by utilizing the loss logits of the rationales.
(IV) the DeClarE~\citep{DBLP:conf/emnlp/PopatMYW18} and Transformer-XH~\citep{DBLP:conf/iclr/ZhaoXRSBT20}, which extract the rationales with the attention score.

\textbf{Setup.}
Our trained GNN-based model adopts the base version of the pre-trained RoBERTa model followed by the GCN with 2 layers,
where each node corresponds to the concatenation of the claim sequence and the textual sequence or cell sequence.
We also insert the \textit{WikiTitle} into the two sequences as the bridge information.
We take the \textit{CLS} token representation as the initial node representations.
The maximum number of input tokens to RoBERTa is 140.
The original model has 85.6\% on label accuracy of claim verification.
The hyperparameters $\lambda_1$, $\lambda_2$, $\lambda_3$ are set to 1, 1, 1 respectively,
and $\lambda_4$, $\lambda_5$ are defined as 5e-3, 0.1 for edge mask, and 0.1, 1 for node mask.
We adopt the instance-level explanation, where each instance is trained 100 epochs, with the learning rate being settled to 1e-2.
For comparison with baselines,
we choose the attention score with the threshold of 0.5 for attention-based methods (DeClarE and Transformer-XH),
the threshold is set to 0.2 for IB since the portion of the gold standard with input is nearly 0.2,
we choose the first two sentences for TSS baseline because of the expensive cost of computing.
All experimental setups of the baselines are followed from the original papers.

\begin{table*}[h]\small
  \renewcommand\arraystretch{1.3}
  \renewcommand\tabcolsep{2.5pt}
  \centering
  \begin{tabular}{cccccccccc}
  \hline
  \multicolumn{2}{c}{\multirow{2}{*}{\textbf{Model}}}                                            & \multicolumn{2}{c}{\textbf{Claim}} & \multicolumn{4}{c}{\textbf{Rationale}}                               & \multicolumn{2}{c}{\textbf{Claim \& Rationale}} \\ \cline{3-10} 
  \multicolumn{2}{c}{}                                                                           & \textbf{F1.c} & \textbf{Acc.c} & \textbf{F1.r} & \textbf{Ext.acc.r} & \textbf{P.r} & \textbf{R.r} & \textbf{Acc.Part}     & \textbf{Acc.Full}    \\ \hline
  \multicolumn{10}{c}{\textbf{Unsupervised}}                                                                                                                                                                                                                   \\ \hline
  \multicolumn{2}{c}{\textbf{TSS-U}}                                                             & 34.61           & 52.93            & 18.75            & 16.83          & 36.57          & 14.59           & 23.77                    & 1.13                 \\
  \multicolumn{2}{c}{\textbf{DeClarE}}                                                           & 68.23           & 69.18            & 27.59            & 13.63          & 31.46          & 31.71           & 43.85                    & 9.81                 \\
  \multicolumn{2}{c}{\textbf{IB-U}}                                                              & 77.30            & 77.30             & 65.28            & 20.08          & 78.01          & 67.30            & 75.36                    & 15.76                \\ \hline
  \multirow{4}{*}{\textbf{\begin{tabular}[c]{@{}c@{}}Edge\\      Mask\end{tabular}}} & \textbf{SaGP}      & \textbf{85.05}$\pm$0.02      & \textbf{85.15}$\pm$0.02       & 80.08$\pm$0.01       & 45.33$\pm$0.05     & 79.15$\pm$0.03     & 88.30$\pm$0.01      & \textbf{82.92}$\pm$0.03      & \textbf{41.17}$\pm$0.05           \\
                                                                                     & -T.       & 85.04$\pm$0.02      & \textbf{85.15}$\pm$0.02       & 80.01$\pm$0.01       & 45.30$\pm$0.06     & 79.14$\pm$0.01     & 88.30$\pm$0.01      & 82.82$\pm$0.03               & 40.11$\pm$0.05           \\
                                                                                     & -C.       & 85.04$\pm$0.05      & 85.15$\pm$0.07       & \textbf{80.25}$\pm$0.16       & \textbf{46.22}$\pm$1.41     & \textbf{79.80}$\pm$0.97     & 87.68$\pm$1.09      & 82.85$\pm$0.06               & 41.14$\pm$1.57           \\
                                                                                     & -T.\&C.   & 85.01$\pm$0.04      & 85.11$\pm$0.04       & 80.15$\pm$0.01       & 45.23$\pm$0.03     & 79.14$\pm$0.01     & \textbf{88.46}$\pm$0.01      & 82.92$\pm$0.05               & 40.01$\pm$0.01           \\ \hline
  \multirow{4}{*}{\textbf{\begin{tabular}[c]{@{}c@{}}Node\\      Mask\end{tabular}}} & \textbf{SaGP}      & 82.24$\pm$0.13      & \textbf{82.26}$\pm$0.13       & 70.47$\pm$0.08       & 38.56$\pm$0.13     & \textbf{75.19}$\pm$0.12     & 76.40$\pm$0.05      & 75.03$\pm$0.08               & 33.61$\pm$0.01           \\
                                                                                     & -T.       & \textbf{82.25}$\pm$0.12      & 82.25$\pm$0.12       & \textbf{70.50}$\pm$0.09       & \textbf{38.60}$\pm$0.10      & \textbf{75.19}$\pm$0.12     & 76.37$\pm$0.07      & 75.04$\pm$0.06      & \textbf{33.65}$\pm$0.04           \\
                                                                                     & -C.       & 81.80$\pm$0.19      & 81.81$\pm$0.19       & 70.34$\pm$0.26       & 36.97$\pm$0.55     & 73.60$\pm$1.05     & \textbf{78.28}$\pm$1.72      & \textbf{75.36}$\pm$0.54               & 32.18$\pm$0.56           \\
                                                                                     & -T.\&C.   & 81.85$\pm$0.15      & 81.85$\pm$0.16       & 70.17$\pm$0.12       & 37.50$\pm$0.18      & 74.27$\pm$0.05     & 77.01$\pm$0.18      & 74.78$\pm$0.22               & 32.64$\pm$0.04           \\ \hline
  \multirow{4}{*}{\textbf{All}}                                                      & \textbf{SaGP}      & \textbf{82.06}$\pm$0.12      & \textbf{82.08}$\pm$0.12       & 70.40$\pm$0.21        & 38.66$\pm$0.27     & 74.99$\pm$0.27     & 76.27$\pm$0.14      & 75.27$\pm$0.81               & 33.90$\pm$0.25           \\
                                                                                     & -T.       & 81.77$\pm$0.11      & 81.78$\pm$0.11       & 70.14$\pm$0.20        & 37.40$\pm$0.36     & 74.23$\pm$0.21     & 76.95$\pm$0.16      & 74.67$\pm$0.15               & 32.66$\pm$0.33           \\
                                                                                     & -C.       & 81.89$\pm$0.09      & 81.90$\pm$0.09       & \textbf{73.64}$\pm$4.80        & \textbf{40.17}$\pm$3.60      & \textbf{75.81}$\pm$2.09     & \textbf{81.11}$\pm$5.70       & \textbf{76.59}$\pm$2.54               & \textbf{34.99}$\pm$3.03           \\
                                                                                     & -T.\&C.   & 82.03$\pm$0.11      & 82.05$\pm$0.11       & 70.38$\pm$0.23       & 38.60$\pm$0.26     & 74.93$\pm$0.28     & 76.30$\pm$0.15      & 74.64$\pm$0.05               & 33.84$\pm$0.24           \\ \hline 
  \multicolumn{10}{c}{\textbf{Supervised}}                                                                                                                                                                                                                     \\ \hline
  \multicolumn{2}{c}{\textbf{BERT Blackbox}}                                                     & 64.72           & 65.20             & -                & -              & -              & -               & -                        & -                    \\
  \multicolumn{2}{c}{\textbf{Pipeline}}                                                          & 69.76           & 69.80             & 77.56            & 44.83          & 76.87          & 86.75           & 62.77                    & 31.23                \\
  \multicolumn{2}{c}{\textbf{TSS-S}}                                                             & 72.99           & 74.36            & 44.15            & 19.42          & 85.67          & 34.12           & 67.75                    & 11.76                \\
  \multicolumn{2}{c}{\textbf{IB-S}}                                                              & 79.14           & 79.17            & 65.68            & 20.08          & 78.91          & 67.31           & 76.70                     & 16.37                \\
  \multicolumn{2}{c}{\textbf{Transformer-XH}}                                                    & 74.05           & 74.33            & 76.70             & 49.10           & 79.43          & 80.47           & 69.17                    & 40.22                \\ \hline
  \multirow{3}{*}{\textbf{SaGP}}                                                     & Edge Mask & \textbf{85.12}$\pm$0.01      & \textbf{85.25}$\pm$0.01       & 80.49$\pm$0.02       & 48.22$\pm$0.01     & 81.18$\pm$0.02     & 86.14$\pm$0.02      & \textbf{82.77}$\pm$0.01               & 43.36$\pm$0.01           \\
                                                                                     & Node Mask & 81.53$\pm$0.06      & 81.54$\pm$0.06       & 84.50$\pm$0.66       & 56.23$\pm$0.23     & 85.51$\pm$0.06     & 86.48$\pm$0.02      & 78.10$\pm$0.11               & 47.67$\pm$0.29           \\
                                                                                     & All     & 82.10$\pm$0.04    & 82.15$\pm$0.03     & \textbf{85.80}$\pm$0.07    & \textbf{61.94}$\pm$0.26     & \textbf{87.89}$\pm$0.07   & \textbf{87.05}$\pm$0.06    & 78.76$\pm$0.11  & \textbf{53.19}$\pm$0.30           \\ \hline
  \end{tabular}
  \caption{\label{tab:results} The performances of different approaches for claim verification and rationale extraction on the FEVEROUS dataset under two settings (mean and standard deviation over three random seed runs), where $-U$, $-S$ denote the unsupervised and supervised version of the model, respectively. $T.$, $C.$ denote the \textit{Topology} and \textit{Compact} properties. $All$ denotes the combination of edge mask and node mask. The best results are marked in bold.}
  \end{table*}

\section{Experimental Results}
\label{sec:experimental_results}

In this section, we evaluate the SaGP model in different aspects. 
Firstly, we compare the overall performance with the baselines using different types of masks under unsupervised and supervised settings. Then, we evaluate the effect of using different diagnostic properties as additional signals for rationale extraction. Finally, following~\citep{DBLP:journals/corr/abs-2102-03322}, we explore various desirable properties of the edge mask.

\subsection{Overall Results}
\label{subsec:overall_results}

\subsubsection{Unsupervised Setting} 
We mainly explore the rationale extraction under the unsupervised setting with different diagnostic properties as the additional signals.
The top of Tab.~\ref{tab:results} reports the overall results of our model against the baselines.
As shown in Tab.~\ref{tab:results}, our model, especially for the edge mask, significantly outperforms the baselines on both the claim verification task and rationale extraction task.
It is worth pointing out that our model outperforms the baselines by over 20\% on the Ext.acc,
which demonstrates the effectiveness of the graph network for extracting rationales.
However, compared with using the edge mask,
the performance of using the node mask decreases on all evaluation metrics in varying degrees,  and nearly 7\% on ACC.Full metrics and 8\% on Ext.acc.
We conjecture that this might attribute to the deficiency of indispensable information in the reasoning process caused by the node mask.
Inherently, the node mask can be considered as the noise signal added to the feature of rationales directly. 

\textbf{Effect of Property.}
We explore the effectiveness of using different diagnostic properties for rationale extraction by conducting the ablation study\footnote{Note that we take the \textit{fidelity} as the primary property for guiding rationale extraction and only conduct the ablation study on the \textit{compact} and \textit{topology} properties.}.
As shown in Tab.~\ref{tab:results},
using \textbf{topology} as an additional objective aims to regularize the graph structure in the multi-hop reasoning process. 
We note a slight decrease when removing the topology from edge mask while no effect occurs for node mask. 
This can be explained by the fact that the topology focuses primarily on the structure other than the input feature.
Furthermore,
for \textbf{compact},
we observe that the standard deviation is relatively higher than other properties on rationale extraction metrics. It might be explained that the extraction process is unstable when perturbing the input.
Although the performance decreases, we still include the compact as a training objective to make the RA-subgraph compact.

\begin{table}[t]\small
  \renewcommand\arraystretch{1.2}
  \renewcommand\tabcolsep{2.5pt}
  \centering
  \begin{tabular}{ccccc}
  \hline
  \multicolumn{2}{c}{\multirow{2}{*}{\textbf{Model}}} & \multicolumn{3}{c}{\textbf{FEVEROUS}}                                \\ \cline{3-5} 
  \multicolumn{2}{c}{}                                & \textbf{Fidelity ($\downarrow$)} & \textbf{Size ($\uparrow$)} & \textbf{Sparsity ($\downarrow$)} \\ \hline
  \multirow{4}{*}{\textbf{SaGP}}          & Edge Mask         & 1.95$\pm$0.59              & 367.40$\pm$0.89         & 3.31$\pm$0.23               \\
                                 & -C.               & 1.53$\pm$0.02              & 361.12$\pm$0.01        & 4.81$\pm$0.22               \\
                                 & -T.                & 1.42$\pm$0.01              & 361.44$\pm$1.24        & 4.88$\pm$0.05              \\
                                 & -C. \& T.          & 1.42$\pm$0.00               & 361.45$\pm$1.21        & 4.80$\pm$0.05              \\ \hline
  \end{tabular}
  \caption{\label{tab:edge_mask} Evaluation of the edge mask matrix. $\downarrow$ denotes the lower is better.}
  \end{table}

\subsubsection{Supervised Setting}
To explore the impact of rationale supervision, 
we replace the compact loss with the rationale label by formulating it as a multi-label problem.
Such rationale supervision can also be considered as the \textit{plausible} property~\citep{DBLP:conf/acl/JainWPW20}.
From the bottom of Tab.~\ref{tab:results} we can see,
our model outperforms all baselines on different metrics of both tasks. 
The ceiling performance of the claim verification task remains the same with the unsupervised setting,
while the rationale supervision brings an improvement on Acc.Full metrics.
Moreover, compared with the unsupervised setting, 
we observe that there is performance improvement in rationale extraction metrics with different types of masks.
It is worth pointing out that the rationale supervision provides more improvement for the node mask compared with the edge mask.
The reason is that the rationale supervision directly forces the model to learn from the rationales while it is not available for edge mask.

\begin{figure}[t]
  \centering
  \includegraphics[scale=0.38]{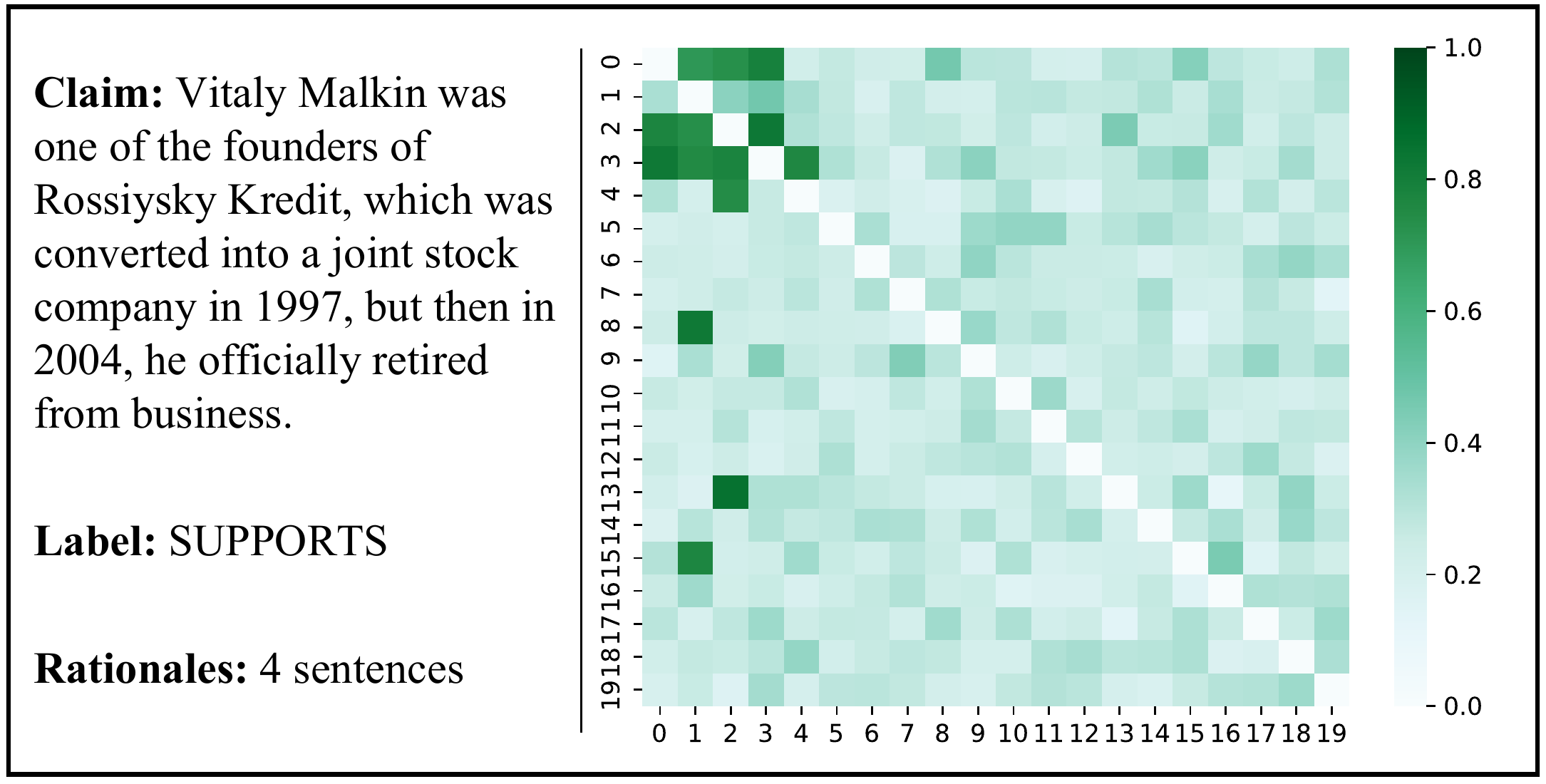}
  \caption{An example with visualization of the edge mask.}
  \label{fig:mask}
\end{figure}

\subsection{Analysis}
\label{subsec:Analysis}
We explore how well the edge mask satisfies the objective of approximating the deletion of the redundant edges.
The node mask is not considered as it mainly focuses on the input features as prior works have done. 
We evaluate the edge mask in terms of three metrics: 
(I) Fidelity measures the proportion of claims where the prediction is retained after the perturbation of the edge.
(II) Size denotes the number of removed edges.
(III) Sparsity measures the proportion of edges that are retained.
As shown in Tab.~\ref{tab:edge_mask},
there are nearly 367.4 edges removed from the total of 420 edges, 
i.e., 3.31\% edges are retained in the graph, while with only 1.95\% decreases in the accuracy of claim. 
This reveals that a large portion of edges in the input graph is not helpful for fact verification, and the model relies heavily on a few significant edges for updating the node representation.
Moreover, we explore the effectiveness of the indirect diagnostic properties of edge mask. 
It can be observed from Tab.~\ref{tab:edge_mask} that 
compared with the compact constrained on the rationale subgraph,
the topology property affects the learning process of edge mask to some extent.
In Fig.~\ref{fig:mask}, 
we visually present the edge mask logits of the example from the test dataset. 
As expected, it clearly shows the significant edges within the adjacency matrix.

\subsection{Error Analysis}
\label{subsec:error_analysis}
\begin{figure}[]
  \centering
  \includegraphics[scale=0.53]{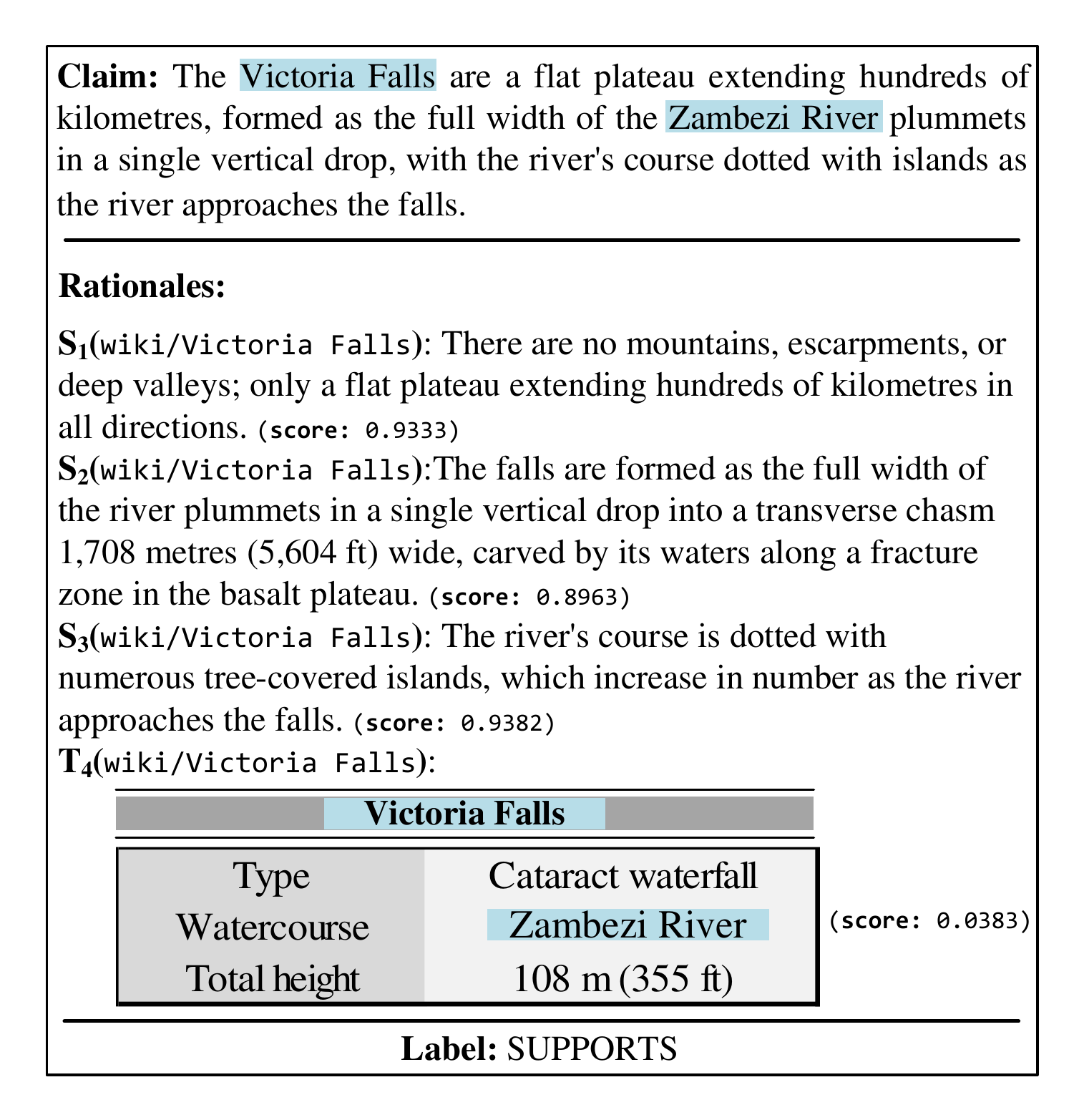}
  \caption{A case with failing to identify rationales within $T4$.}
  \label{fig:case1}
\end{figure}

We conduct an error analysis on rationales extracted by our model on 50 randomly chosen examples from the test set of the FEVEROUS dataset.
The main errors are summarized as follows:
(I) the noisy evidence contains bags of tokens that highly overlap with the claim, which brings difficulty in accurately understanding the semantic information for the model.
(II) the model fails to capture the implicit correlation between different entities, especially for the table cells.
For example, in Fig.~\ref{fig:case1}, the relation between ``\textit{Victoria Falls}'' and ``\textit{Zambezi River}''  present in the table while not mentioned in the sentence.
This scenario requires the model with a higher-level understanding to parse the structure of tables.

\section{Conclusion}
\label{sec:conclusion}
By framing explainable multi-hop fact verification as the subgraph extraction, we propose a novel GCN-based method with salience-aware graph learning to jointly model the multi-hop fact verification and the rationale extraction. Moreover, we introduce three diagnostic properties as additional training objectives to improve the quality of the extracted rationale in the multi-task model. The results on the FEVEROUS benchmark dataset demonstrate the effectiveness of our model. In the future, we will explore how to adapt the model to other domains.

\section{Acknowledgments}
We would like to thank anonymous reviewers for their valuable comments and helpful suggestions.
We would also like to thank Professor Yulan He for her valuable suggestions for our paper.
This work was funded by the National Natural Science Foundation of China (62176053).

\bibliography{custom}

\end{document}